\documentclass[letterpaper, 10 pt, journal, twoside]{ieeetran}  
\IEEEoverridecommandlockouts                              
 
\usepackage[T1]{fontenc}
\usepackage{mathptmx} 
\usepackage{times} 
\usepackage{amsmath} 
\usepackage{amssymb}  
\usepackage{graphicx} 
\usepackage[ruled,vlined]{algorithm2e} 
\usepackage{booktabs}  
\usepackage{float}
\usepackage{cite}
\usepackage{times} 
\usepackage{amssymb}  
\usepackage{bigstrut}
\usepackage{multirow}
\usepackage{balance}
\usepackage[hidelinks=false,bookmarks=false]{hyperref}  
\usepackage{multirow}
\usepackage{lineno}
\usepackage{amsopn}
\usepackage{comment}
\usepackage{color}
\usepackage{tabularx} 
\usepackage{subfigure}
\usepackage{url}
\usepackage{hyperref}    
\usepackage{bm}
\usepackage{balance}

\DeclareMathAlphabet{\mathcal}{OMS}{cmsy}{b}{n}
\DeclareMathAlphabet{\mathcal}{OMS}{cmsy}{m}{n}
\usepackage{balance}
 
\title{The Role of the Hercules Autonomous Vehicle During the COVID-19 Pandemic: An Autonomous Logistic Vehicle for Contactless Goods Transportation}  
\author{Tianyu Liu$^{*}$, Qinghai Liao$^{*}$, Lu Gan, Fulong Ma, Jie Cheng, Xupeng Xie, Zhe Wang, Yingbing Chen, Yilong Zhu, Shuyang Zhang, Zhengyong Chen, Yang Liu, Meng Xie, Yang Yu, Zitong Guo, Guang Li, Peidong Yuan, Dong Han, Yuying Chen, Haoyang Ye, Jianhao Jiao, Peng Yun, Zhenhua Xu, Hengli Wang, Huaiyang Huang, Sukai Wang, Peide Cai, Yuxiang Sun, Yandong Liu, Lujia Wang, and Ming Liu 
\thanks{$^{*}$ Equal Contribution}
}
\IEEEaftertitletext{\vspace{-2.9\baselineskip}}

\begin{document} 

\maketitle 

Since early 2020, the coronavirus disease 2019 (COVID-19) has spread rapidly across the world. As at the date of writing this article, the disease has been globally reported in 223 countries and regions, infected over 108 million people and caused over 2.4 million deaths\footnote{\url{https://covid19.who.int/} (accessed on Feb. 17, 2021)}. Avoiding person-to-person transmission is an effective approach to control and prevent the pandemic. However, many daily activities, such as transporting goods in our daily life, inevitably involve person-to-person contact. Using an autonomous logistic vehicle to achieve contact-less goods transportation could alleviate this issue. For example, it can reduce the risk of virus transmission between the driver and customers. Moreover, many countries have imposed tough lockdown measures to reduce the virus transmission (e.g., retail, catering) during the pandemic, which causes inconveniences for human daily life. Autonomous vehicle can deliver the goods bought by humans, so that humans can get the goods without going out. These demands motivate us to develop an autonomous vehicle, named as Hercules, for contact-less goods transportation during the COVID-19 pandemic. The vehicle is evaluated through real-world delivering tasks under various traffic conditions. 

\begin{figure}[t] 
\setlength{\abovecaptionskip}{0pt} 
\setlength{\belowcaptionskip}{0pt} 
\centering
\includegraphics[width=1.0\linewidth]{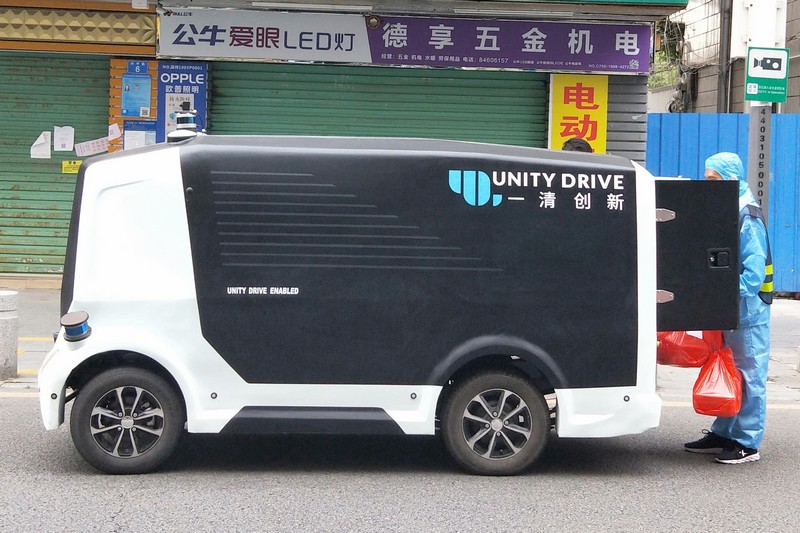} 
\caption{A worker dressed in a protective suit is collecting goods from our Hercules logistic autonomous vehicle. There is no person-to-person contact during the process of goods transportation. This photo was taken in Shenzhen, Guangdong, China, February 2020 during the COVID-19 pandemic.} 
\label{fig_hercules_vehicle} 
\end{figure}
 
  
\begin{figure}[t] 
\setlength{\abovecaptionskip}{0pt} 
\setlength{\belowcaptionskip}{0pt} 
\centering 
\raggedleft
\subfigure[The sensors on the vehicle (with the cargo box removed).]{
\includegraphics[width=1.0\linewidth]{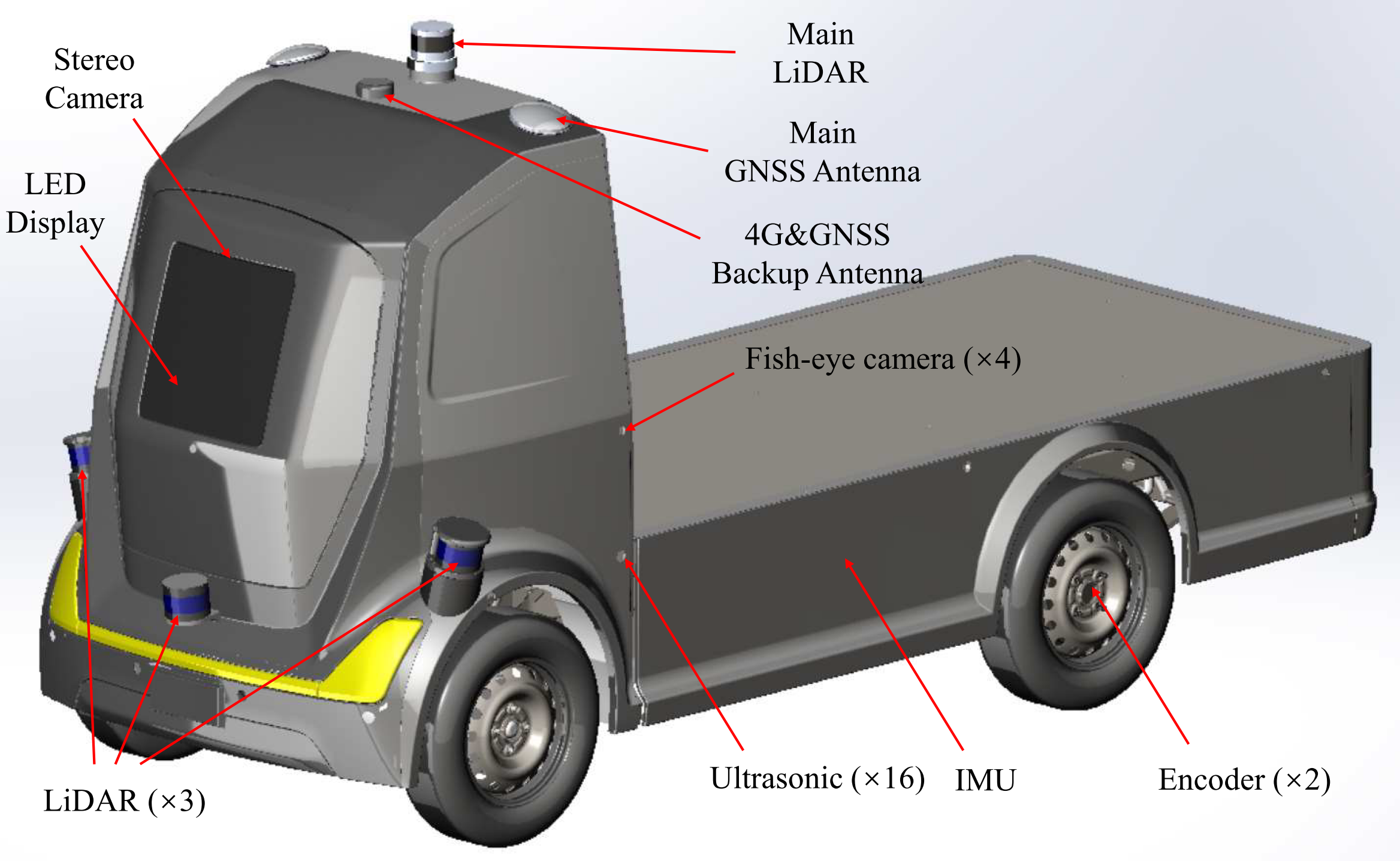} 
\label{fig_hardware_on_3d_model} 
}  
\raggedleft
\subfigure[The sensors and control units on the mobile base (chassis).]{
\includegraphics[width=1.0\linewidth]{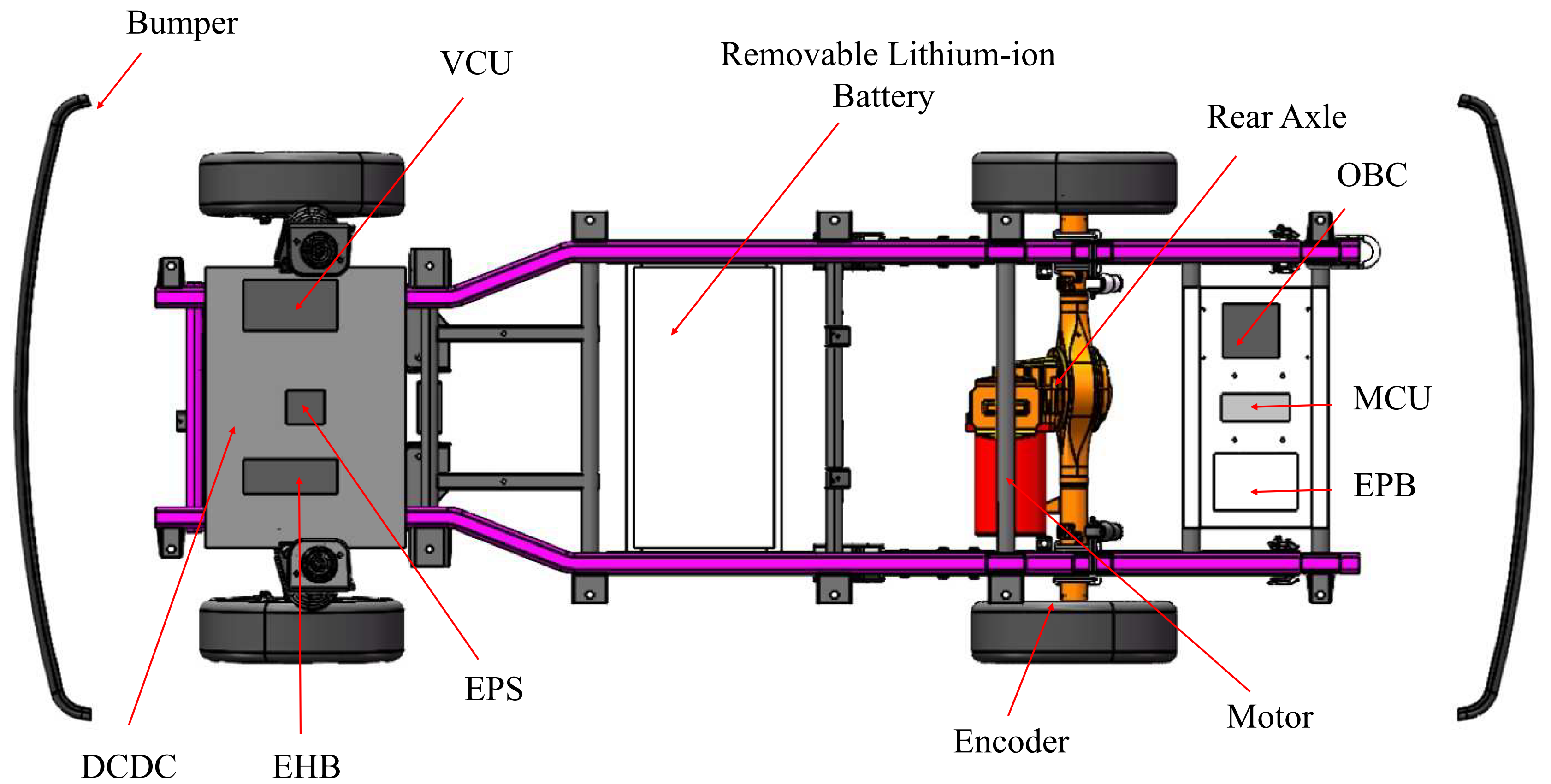} 
\label{fig_hardware_mobile_base}
} 
\caption{The sensors used in our vehicle and the modules in the chassis. Note that the cargo box is replaceable. It is not shown in the figure.}
\label{fig_hardware} 
\end{figure}

\begin{figure*}[t]
\setlength{\abovecaptionskip}{0pt} 
\setlength{\belowcaptionskip}{0pt} 
\centering
\includegraphics[scale=0.222]{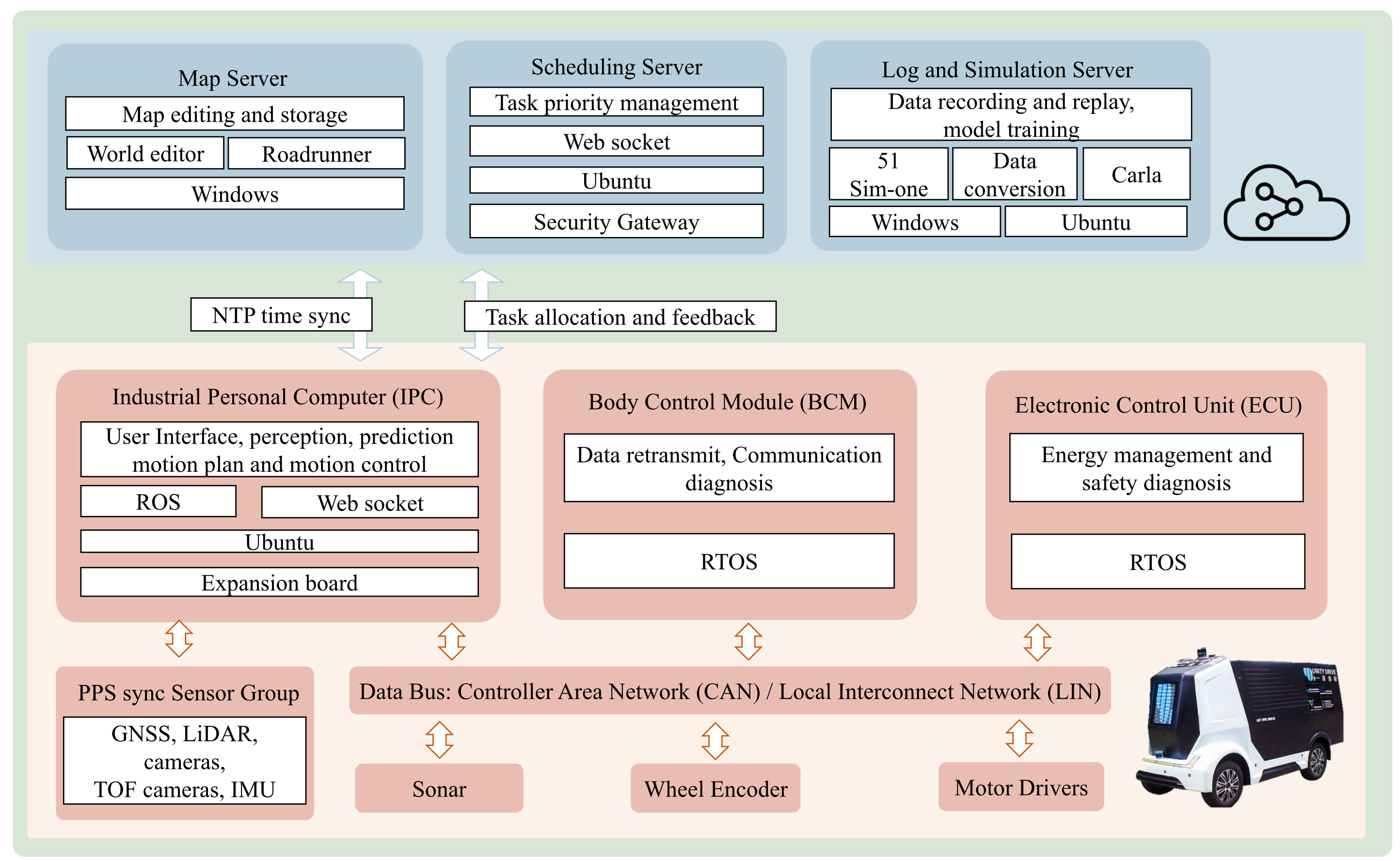}  
\caption{The software architecture of our vehicle. The part shown in the yellow box is running on the vehicle, and the part shown in the blue box is running on the cloud. The figure is best viewed in color.}
\label{fig_software_architecture}
\end{figure*}

There exist many studies related to autonomous vehicles, however, most of these works focus on the specific modules of autonomous driving systems. For example, Sadigh \textit{et al.} \cite{sadigh2016planning} developed a planning method that models the interaction with other vehicles. Koopman \textit{et al.} \cite{koopman2016challenges} presented a testing paradigm for autonomous vehicles. 
Some researchers have tried to construct the complete autonomous driving systems \cite{endsley2017autonomous}. Compared with these studies, we build a complete system and add several new modules, such as the cloud server module. We also make some adjustments, such as considering novel dynamic constraints, in our solution to make the vehicle more suitable for the contact-less good transportation. In the following sections, we provide details on the hardware, software, as well as the algorithms to achieve autonomous navigation including perception, planning and control. This paper is accompanied by a demonstration video and a dataset, which are available at \url{https://sites.google.com/view/hercules-vehicle}.

\section{Hardware System}
 
The hardware system of our vehicle mainly consists of a fully functional Drive-by-Wire (DBW) chassis and autonomous driving-related devices. Fig. \ref{fig_hardware} shows the sensors and the 3-D model of our vehicle.

\subsection{Fully Functional DBW Chassis}
To achieve autonomous driving, the first task is to equip a vehicle with the full Drive-by-Wire (DBW) capability. Our DBW chassis can be divided into four parts: 1) Motion control. This module includes Motor Control Unit (MCU), Electric Power Steering (EPS), Electro-Hydraulic Brake (EHB) and Electronic Parking Brake (EPB). MCU supports both the speed control and torque control. EPS controls the steering angle and speed of the the vehicle. The combination of MCU and EPS controls the longitudinal and lateral motions. EHB controls the brake. EPB controls the parking brake; 2) Electronic accessories control. A vehicle has some basic electronic accessories such as lights and horns. They are controlled by the Body Control Module (BCM); 3) Basic sensors. Our chassis are equipped with some basic sensors, such as bumper, wheel encoder and Tire Pressure Monitoring System (TPMS). The bumper and TPMS are both safety-critical sensors. Specifically, bumper is used to detect collisions and is the last defence to prevent further damage when accident occurs; 4) System control. The chassis system is controlled and managed via Vehicle Control Unit (VCU) which is responsible for coordinating each module. It keeps communicating with the Industrial Personal Computer (IPC), performing parameter checking and sending commands to other modules. In addition, VCU is responsible for critical safety functions, such as the stopping signal from the emergency button. In our chassis, VCU and BCM are implemented on one device.

There are two batteries in our vehicle. A 12 V Lead-acid starter battery and a 72 V removable lithium-ion battery, which can support the maximum 80 Km running distance. The lithium-ion battery powers the chassis, IPC, sensors and accessories. It has a built-in Battery Management System (BMS) to monitor and manage the battery. The removable design allows the vehicle to operate at 24 hours a day without stopping for a recharge. An On-Board Charger (OBC) with a Direct Current Direct Current (DCDC) converter takes about 5 hours to fully charge the battery.  

\subsection{Autonomous Driving-related Devices}
The devices related to autonomous driving are: 
1) Computation platform. Our vehicle is equipped with an IPC which has an Intel i7-8700 CPU with 6 cores and 12 threads, 32 GB memory, and a 1050Ti NVIDIA Graphics card. It is able to run deep learning-based algorithms; 
2) Sensors. As shown in Fig. \ref{fig_hardware_on_3d_model}, our vehicle is equipped with four 16-beam LiDAR, one MEMS short-range LiDAR, four fish-eye cameras, $4\times4$ ultrasonic radars, one IMU and one high-precision GNSS system which supports RTK and heading vector;
3) Auxiliary devices. We have 4G/5G Data Transfer Unit (DTU), Human Machine Interface (HMI), LED display, remote controller. The DTU allows the IPC to be connected to our cloud management platform via the Internet. The LED display is programmable and can be controlled by the IPC. Hence, it can interact with other traffic participants like pedestrians and human drivers. Also, it can be used for advertisement. The remote controller is necessary in the current stage to ensure safety.
  
\section{Software System} 

Fig. \ref{fig_software_architecture} shows the software architecture of our autonomous vehicle. It can be generally divided into two parts: software system running on the vehicle, and software system running on the cloud. 

\subsection{Software System on the Vehicle}
There are three main computing platforms on the vehicle: the IPC, Electronic Control Unit (ECU) and BCM. The IPC is used to run algorithms for autonomous navigation. The ECU is used to ensure the safety of the vehicle by energy management and safety diagnosis. 
The applications on ECU run on the Real-Time Operating System (RTOS), which satisfies the real-time requirements. 
The BCM connects the IPC and ECU. It also runs on the RTOS, which meets the real-time requirements. It detects the communication between the nodes of the CAN network by heartbeat protocols. When major nodes of this network experience an outage or crash, the BCM stops transmitting high-level CAN signals from the IPC to the VCU and waits for human interventions.  

The sensors on the vehicle are synchronized by a 1 Hz Pulse per Second (PPS) signal from the external GNSS receiver. The IPC receives data from the sensors and uses them at different frequencies. For example, the state estimation module updates at 100 Hz, providing real-time enough position feedback for the control system. This is achieved by fusing LiDAR measurements with other high-frequency sensors, e.g., IMU or wheeled odometer. The LiDAR object detection module runs at 10 Hz according to the refresh rate of the LiDAR. All the modules are developed on the Robot Operating System (ROS) to facilitate data communication.

\begin{figure}[t]
\setlength{\abovecaptionskip}{0pt} 
\setlength{\belowcaptionskip}{0pt} 
\centering
\includegraphics[width=1.0\columnwidth]{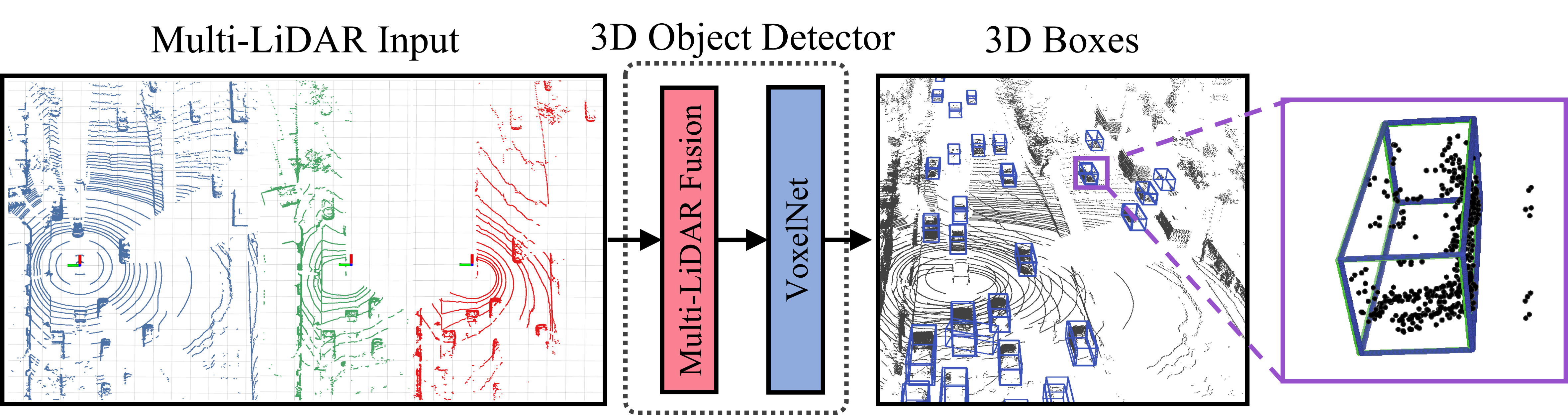}
\caption{Overview of the 3-D object detection module. The inputs are multiple point clouds captured by synchronized and well-calibrated LiDARs. We use an early-fusion scheme to fuse the data from multiple calibrated LiDARs, and adopt the VoxelNet \cite{zhou2018voxelnet} to detect 3-D objects from the fusion results.}
\label{fig_object_detection}
\end{figure}

\subsection{Software System on the Cloud}

The software system on the cloud mainly includes the map server, the scheduling server and the log and simulation server. The map server stores pre-built maps. The scheduling server performs the task allocations and collects the status of every registered running vehicle. It also plays the role of accessing the map data for routing, transmitting sensor data into the map server, recording the key information into the log server, and replaying the data recorded for a good trace-back. The log and simulation server run the end-to-end simulator Carla and 51Sim-One. The clock synchronization between the platforms on the vehicle and cloud is manipulated based on the network time through the Network Time Protocol (NTP). 

\section{Perception}
Perception serves as the fundamental component of autonomous navigation. It provides necessary information for planning and control. This section describes two key perception technologies used in our vehicle.

\subsection{Multiple Lidar-based 3-D Object Detection}
 
The 3-D object detection aims to recognize and classify objects, as well as estimate their poses with respect to a specific coordinate system. We use multiple Lidars for object detection. The first step is to calibrate the Lidars. In this work, we propose a marker-based approach \cite{jiao2019novel} for automatic calibration without any additional sensors and human intervention. We assume that three linearly independent planar surfaces forming a wall corner shape are provided as the calibration targets, ensuring that the geometric constraints are sufficient to calibrate each pair of Lidars. After matching the corresponding planar surfaces, our method can successfully recover the unknown extrinsic parameters with two steps: a closed-form solution for initialization based on the Kabsch algorithm \cite{kabsch1978discussion} and a plane-to-plane iterative closest point (ICP) for refinement. 
The overview of our 3-D object detection is shown in Fig. \ref{fig_object_detection}. The inputs to our approach are multiple point clouds captured by different Lidars. We adopt an early-fusion scheme to fuse the data from multiple calibrated Lidars at the input stage. With the assumption that the Lidars are synchronized, we transform the raw point clouds captured by all the Lidars into the base frame, and then feed the fused point clouds into the 3-D object detector \cite{zhou2018voxelnet}. The final output is a series of 3-D bounding boxes. 

\begin{figure}[t]
\setlength{\abovecaptionskip}{0pt} 
\setlength{\belowcaptionskip}{0pt} 
\centering
\includegraphics[width=0.485\textwidth]{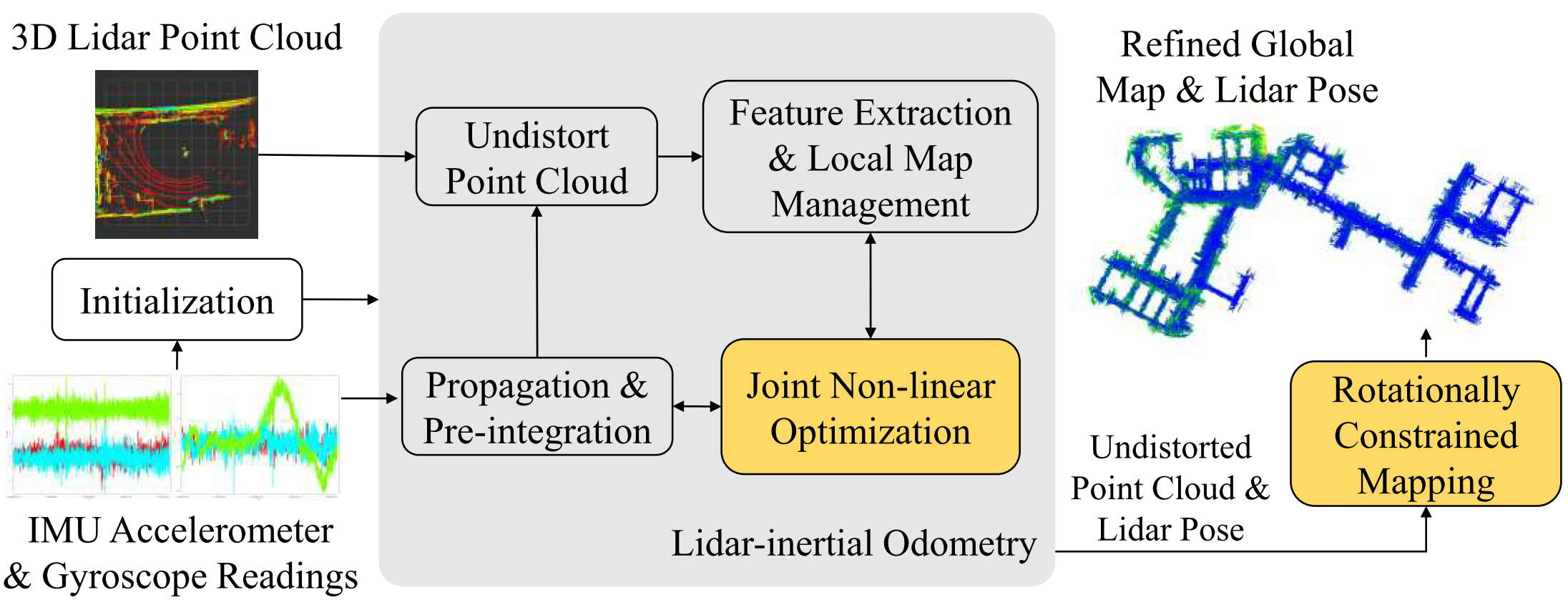}
\caption{The schematic diagram of our 3-D point-cloud mapping system. After the initialization, the system will estimate the states and refine the global map and Lidar poses in the odometry and mapping sub-modules, respectively.}
\label{fig_mapping_system}
\end{figure}
 
\begin{figure*}[t]
\setlength{\abovecaptionskip}{0pt} 
\setlength{\belowcaptionskip}{0pt} 
\centering
\includegraphics[width=2.0\columnwidth]{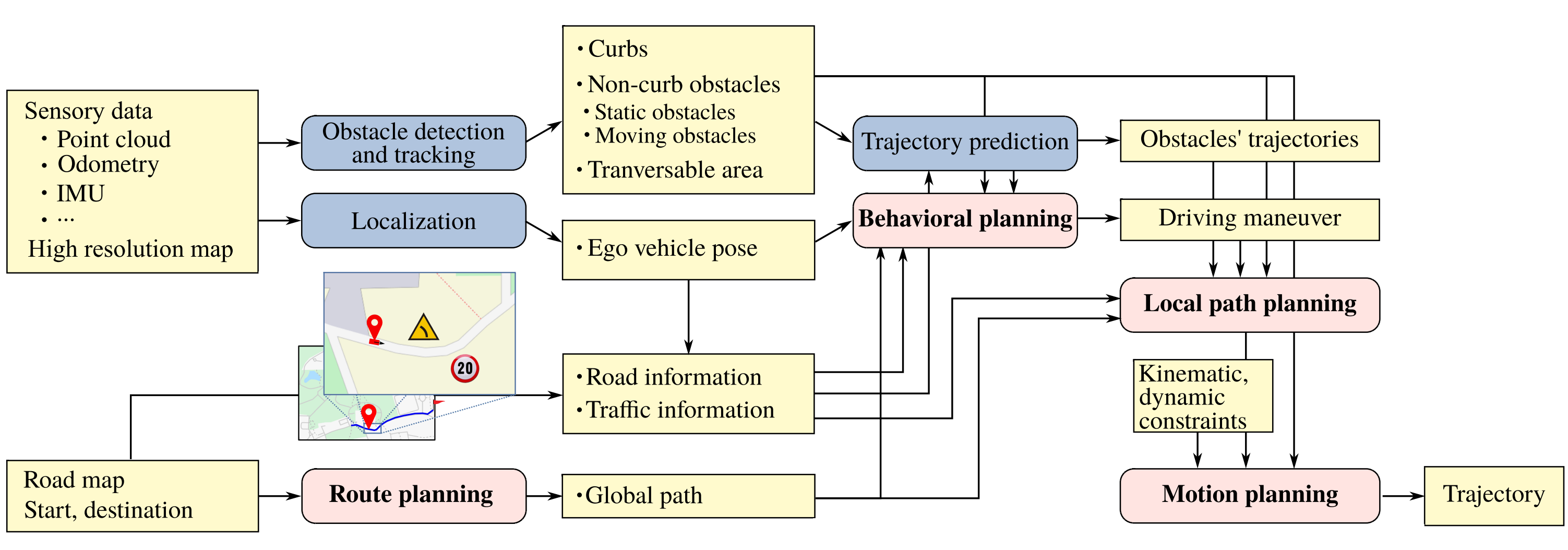}
\caption{The schematic diagram of planning for our autonomous vehicle. The planning process consists of four layers: route planning, behavioral planning, path planning and motion planning. The four layers are coloured in pink. }
\label{fig:planning}
\end{figure*}

\subsection{3-D Point-cloud Mapping} 

The 3-D point-cloud mapping aims to build the 3-D map of the traversed environments. Fig. \ref{fig_mapping_system} shows the diagram of our mapping system. The inputs to the system are the raw data from the IMU and 3-D Lidar (i.e., accelerometer and gyroscope readings from the IMU and point clouds from the 3-D Lidar). The system starts with an adapted initialization procedure, followed by two major sub-modules: the Lidar-inertial odometry and rotationally constrained mapping. 
Since the vehicle usually remains still at the beginning of mapping, we do not need to excite the IMU to initialize the module as described in \cite{ye2019tightly}, which is more suitable for hand-held applications. With the stationary IMU readings, the initial orientation for the first body frame can be obtained by aligning the average of the IMU accelerations to the opposite of the gravity direction in the world frame. The initial velocity, and IMU biases are set to zero. Then, the Lidar-inertial odometry optimally fuses Lidar and IMU measurements in a local window. The mapping with rotational constraints further refines the Lidar poses and the point-cloud map.
 
\section{Planning}
Planning enables autonomous vehicles to acquire future paths or motions towards the destination. The planning for autonomous driving is challenging, because traffic environments are usually with dynamic objects, bringing about risks and uncertainties \cite{liu2015robotic}.   
Autonomous vehicles are required to interact with various road participants, including cars, motorcycles, bicycles, pedestrians, etc. The planner needs to meet the vital requirements of safety, and the kinematic and dynamic constraints of vehicles, as well as the traffic rules. To satisfy these requirements, our planning is hierarchically divided into four layers: route planning, behavioral planning, local path planning, and motion planning. Fig. \ref{fig:planning} shows the four-layer planning process. 

\subsection{Route Planning}
The route planning aims at finding the global path from the global map. For autonomous driving, the route planner typically plans a route given a road network. For structured environments with clear road maps, we use path planning algorithm A* to find the route by establishing the topological graph. However, driveways in industrial parks or residential areas are often not registered in the road net. Furthermore, some of the traversable areas in these places are unstructured and not clearly defined. We employ experienced drivers as teachers to demonstrate reference routes in these places. Fig.\ref{fig:Impression} shows the global routes in the road network with arrows indicating the forward directions.
 
\subsection{Behavioral Planning}
Behavioral planning decides the manoeuvres for local navigation. It is a high-level representation of a sequence of vehicle motions. Typical manoeuvres are lane keeping and overtaking. This layer receives information from the global maps and finds the category of the local area to give specifications on path planning. For example, unstructured environments, like parking lots, have different requirements on planning. Given the road map and the localization of the ego-vehicle, features of the local area can be obtained. As shown in Fig. \ref{fig:planning}, road information that indicates the road environment classification of the global path segments is helpful for behavioral planning. 
Furthermore, traffic information from traffic signs helps in making decisions. The road and traffic information together with the estimation of other moving agents allows the behavioral planner to follow or overtake the front car, or pull over the ego-vehicle.   

\subsection{Local Path Planning}
The local path planning generates a geometric path from the starting pose to the goal pose for the vehicle to follow. The time complexity of this process increases with increased path length, so it is often limited to a local range to ensure real-time planning. The local path planner needs to tackle motion constraints of the vehicle to generate collision-free paths that conform to the lane boundaries and traffic rules.  
Fig. \ref{fig:planning} shows the online local path planning for driving on standard roads. Here we plan the path in the Frenet coordinate system. With the global path as the reference path, it defines the lateral shift to the path and the distance travelled along the path from the start position. We drew multiple samples with different speeds and lateral offsets. Then a graph search method is adopted to search the path with the minimum cost. To define the cost of the coordinates of each curve, we take into consideration the quality of the curve, ending offset to the global path, and other factors (e.g., the potential trajectories of other agents).
  
\subsection{Motion Planning} 
Given the planned path, motion planning is the final layer which optimizes the trajectory with dynamic constraints from the vehicle, the requirements for comfort and energy consumption. The planned trajectory specifies the velocity and acceleration of the vehicle at different timestamps, so it is also called trajectory planning. 
Though the path planned in the Frenet frame contains speed information, the dynamic constraint of the vehicle is not yet considered. Besides this, the local planning process is time-consuming and has a low update rate, which is inadequate to handle dynamic obstacles and emergency cases. The motion planner optimizes the trajectory given the information of obstacles, the constraints from the vehicle, and the path from the local path planner. It outputs the final trajectories for the controller to follow at a much higher updating rate to ensure safety.

\section{Control}

\begin{figure}
\setlength{\abovecaptionskip}{0pt} 
\setlength{\belowcaptionskip}{0pt} 
\centering
\includegraphics[width=1.0\columnwidth]{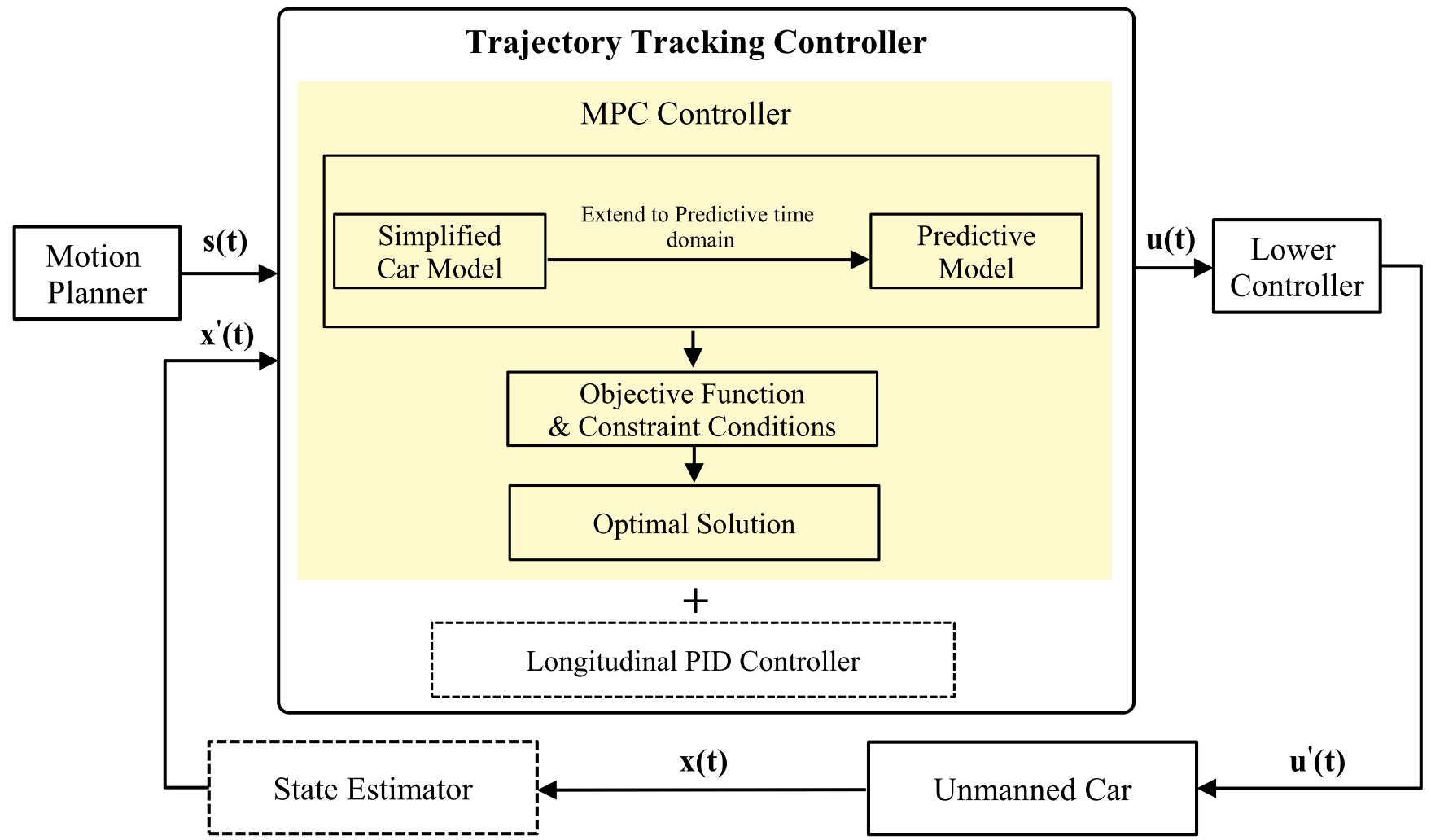}
\caption{The schematic diagram of our controller. The main component is the trajectory tracking controller.}
\label{fig_control}
\end{figure}

The main task of vehicle control is to track the planned trajectory. In the past decade, many trajectory tracking controllers have been developed, among which the Model Predictive Controller (MPC) \cite{garcia1989model} is the most popular one. The schematic diagram of our controller is shown in Fig. \ref{fig_control}. 
 
As we can see, there are two inputs to the trajectory tracking controller. One is the trajectory $s(t)$, which includes the information (e.g., desired coordinates, curvatures, speed) from the motion planner, the other is the feedback information $x'(t)$ from the state estimator. Sometimes, sensor feedback from the chassis cannot be directly sent to the controller, or more feedback quantities are required by the controller, which is difficult to obtain from sensors. In such cases, a state feedback estimator is required but not a must. In Fig. \ref{fig_control}, the output of the trajectory tracking controller $u(t)$ is sent to the chassis after being processed by a lower controller. The lower controller can work for many purposes. For example, our autonomous vehicle can work in both the autonomous-driving mode and the parallel-driving model (i.e., the remote control mode). The trajectory tracking controller only functions in the autonomous-driving mode, which means that only in this mode the lower controller takes as input $u(t)$. 

The vehicle control can be divided into the lateral control, which controls steer angles, and the longitudinal control, which controls the car speed. There are two types of MPCs in the area of the autonomous vehicle. One is kinematics-based while the other is dynamics-based. The kinematics-based MPC is a combined controller that integrates the lateral control and longitudinal control. Therefore, the longitudinal PID controller highlighted in the dashed box in Fig. \ref{fig_control} may be not required. The vector of two control quantities $u(t)$ (i.e., steer angle and speed), will be directly given by the MPC. However, the dynamics-based MPC is a standalone lateral controller of which the output is a control quantity of the steering angle. In such a case, a longitudinal PID controller that outputs the speed control quantity will be required. And the outputs of these two controllers constitute $u(t)$. 

\begin{figure*}
	\begin{minipage}[b]{0.33\linewidth}
		\subfigure[]{\includegraphics[width=1\linewidth]{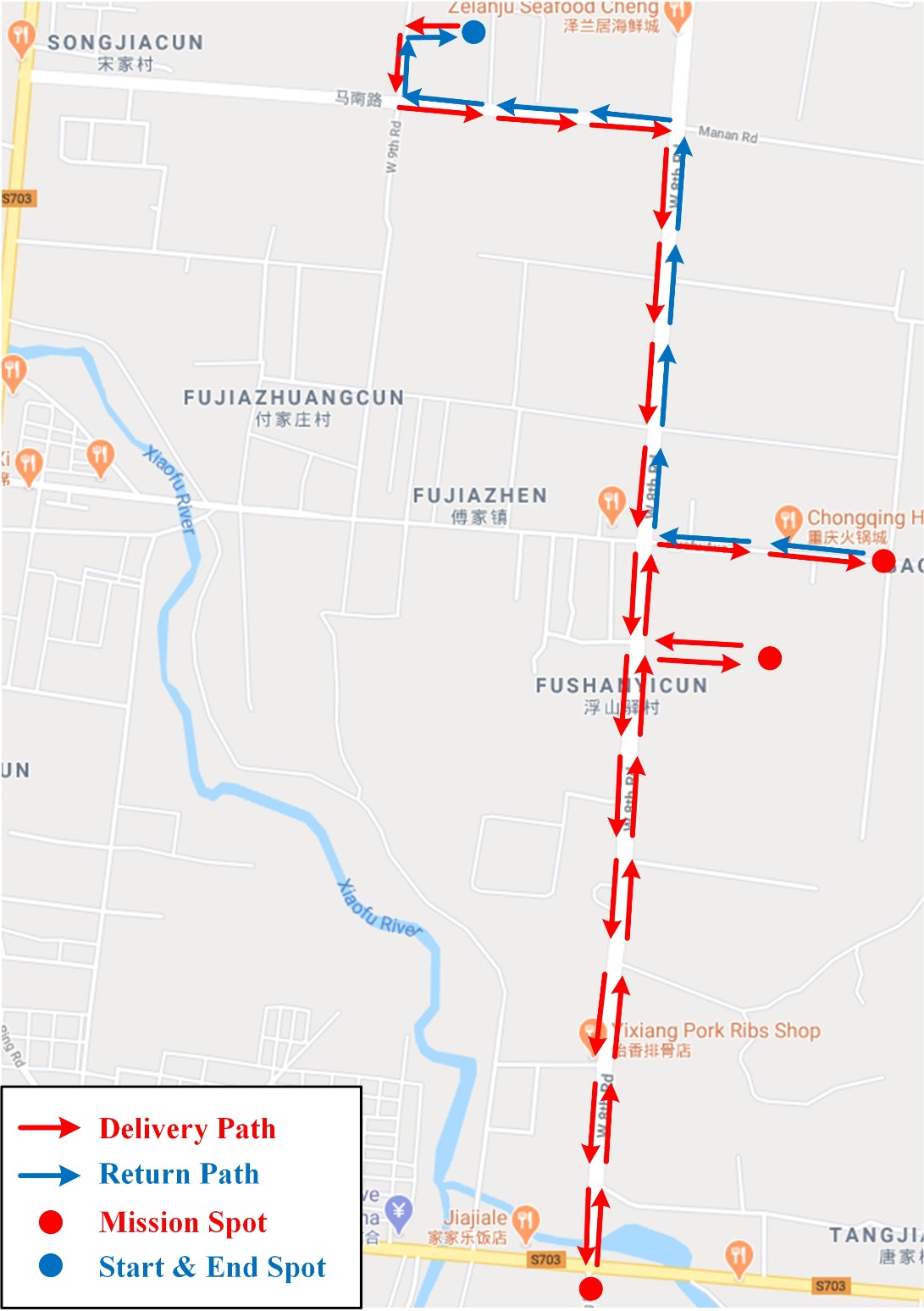} }
	\end{minipage}\hspace{2.5pt}\begin{minipage}[b]{0.66\linewidth}
		\subfigure[]{\includegraphics[width=0.5\linewidth]{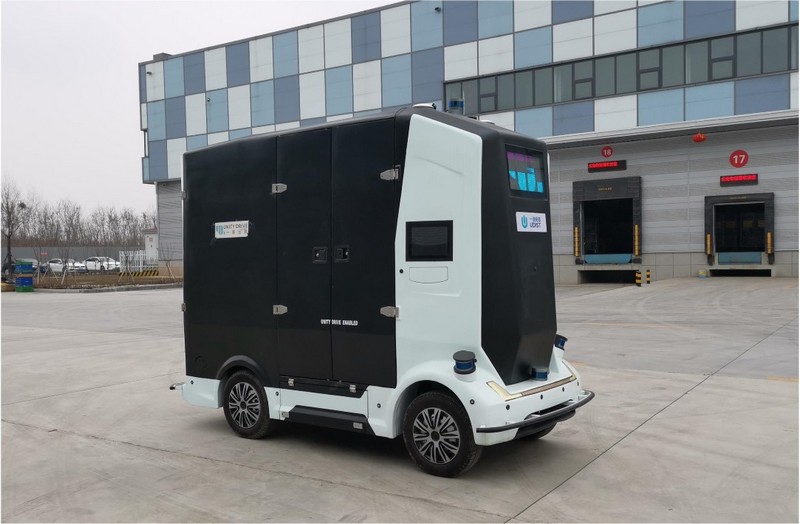}}\hspace{2.5pt}\subfigure[]{\includegraphics[width=0.5\linewidth]{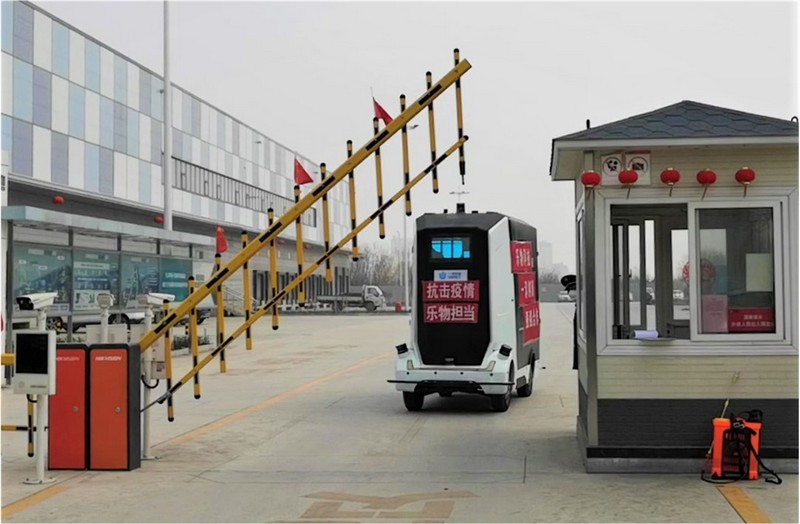}} \vspace{-2mm} \\
		\subfigure[]{\includegraphics[width=0.5\linewidth]{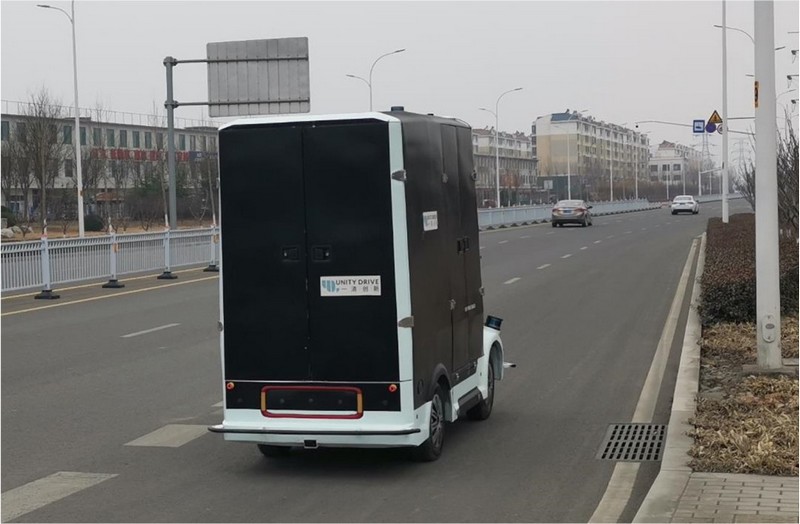}}\hspace{2.5pt}\subfigure[]{\includegraphics[width=0.5\linewidth]{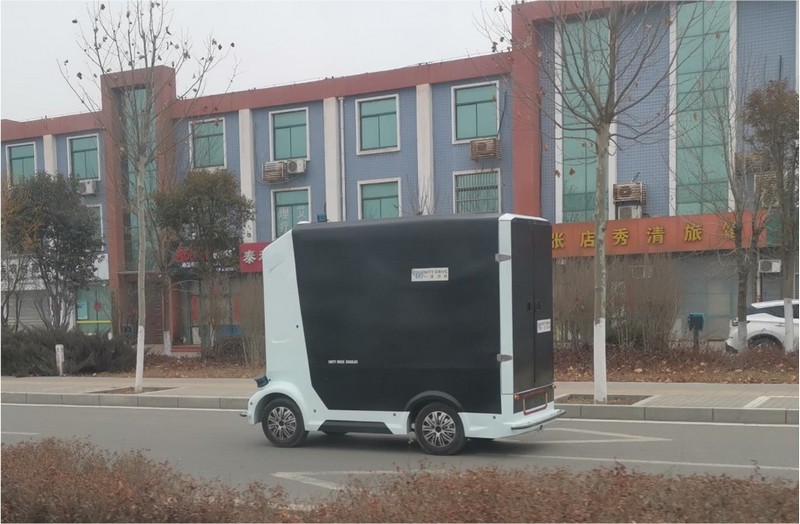}}
	\end{minipage}
	\vspace{-2mm} \\
	\begin{minipage}[b]{1\linewidth}
		\subfigure[]{\includegraphics[width=0.33\linewidth]{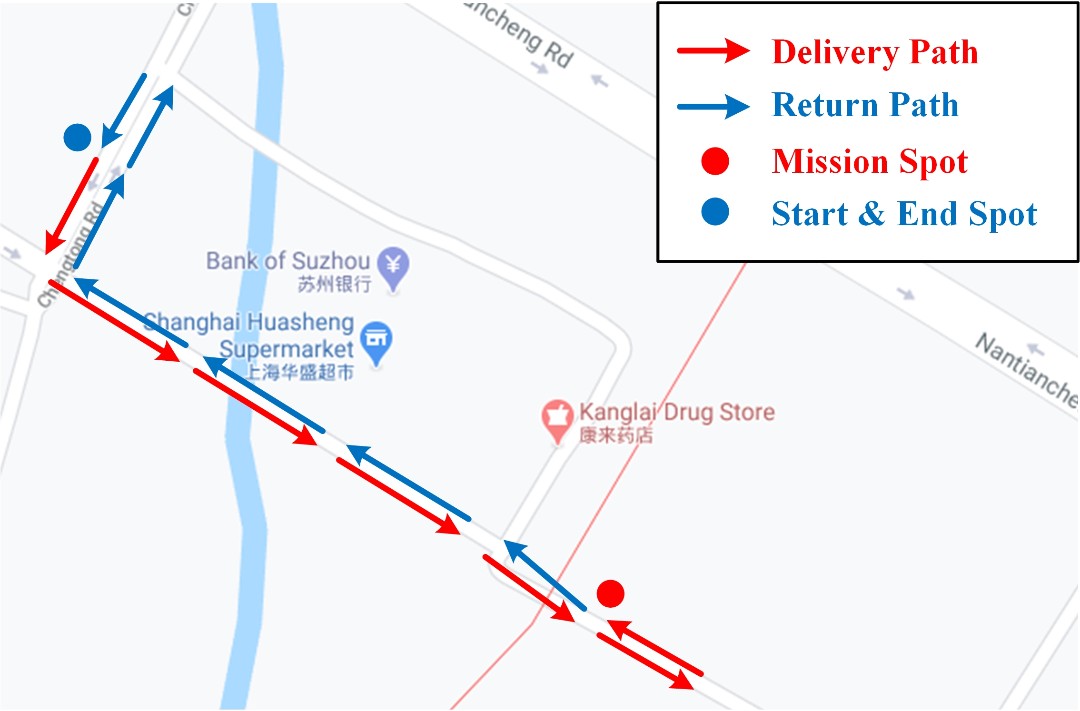}}\hspace{2.5pt}\subfigure[]{\includegraphics[width=0.33\linewidth]{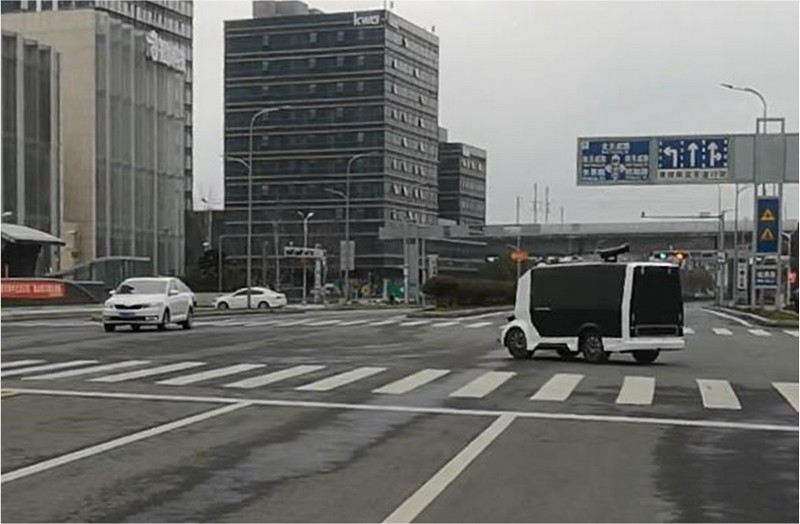}}\hspace{2.5pt}\subfigure[]{\includegraphics[width=0.33\linewidth]{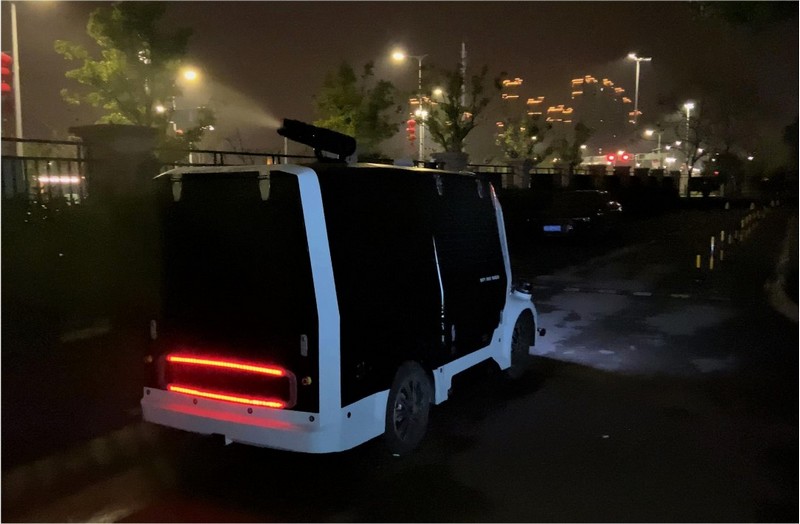}}
	\end{minipage}
	\vspace{-2mm} \\
	\begin{minipage}[b]{1\linewidth}
		\subfigure[]{\includegraphics[width=0.33\linewidth]{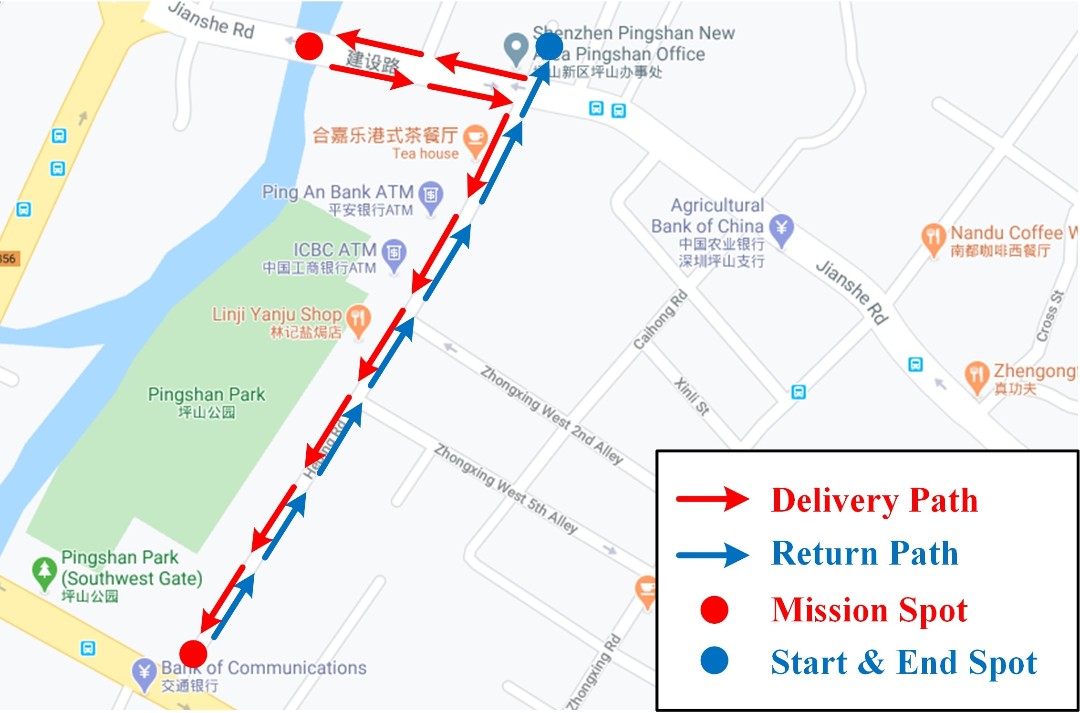}}\hspace{2.5pt}\subfigure[]{\includegraphics[width=0.33\linewidth]{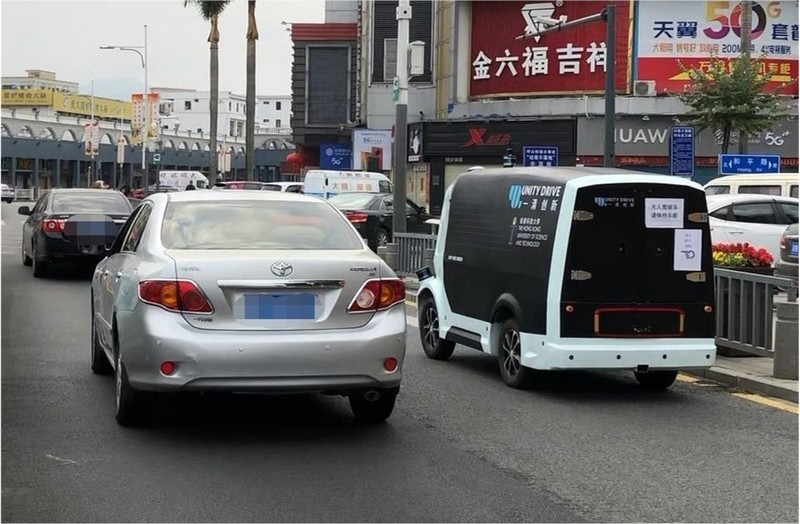}}\hspace{2.5pt}\subfigure[]{\includegraphics[width=0.33\linewidth]{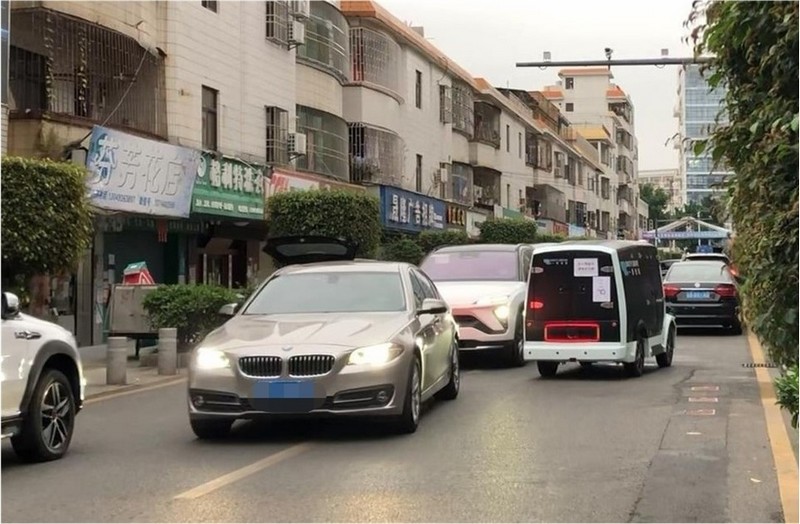}}
	\end{minipage}
	\vspace{-2mm} \\
	\begin{minipage}[b]{1\linewidth}
		\subfigure[]{\includegraphics[width=0.33\linewidth]{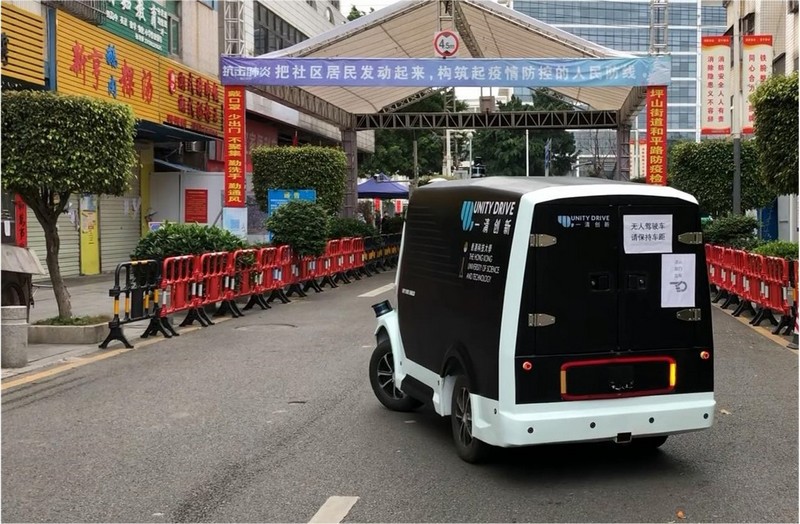}}\hspace{2.5pt}\subfigure[]{\includegraphics[width=0.33\linewidth]{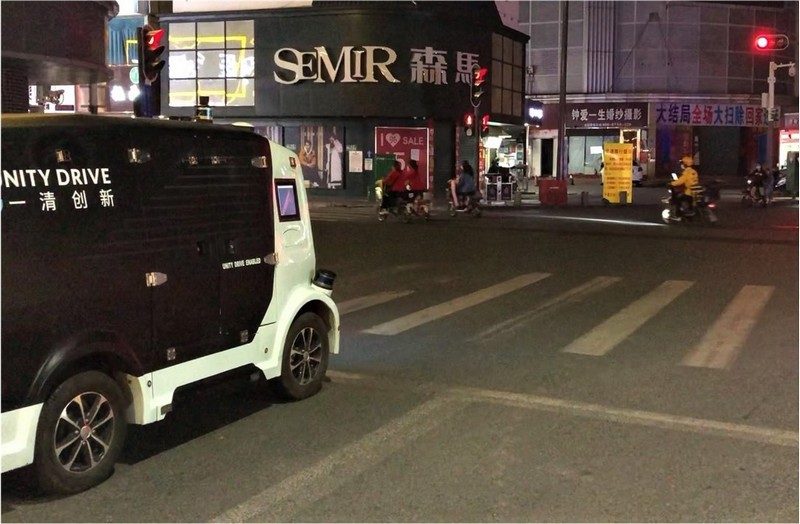}}\hspace{2.5pt}\subfigure[]{\includegraphics[width=0.33\linewidth]{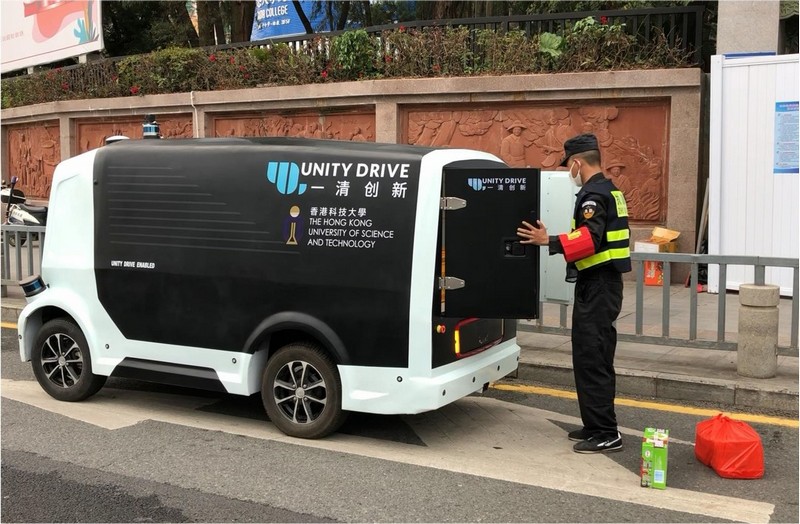}}
	\end{minipage}
	\caption{Selected task routes and demonstration photos. The first column includes three representative routes: (a) A 9.6 Km route in Zibo, Shandong for vegetable delivery; (f) A 1.2 Km route in Suzhou, Jiangsu for lunch meal delivery; (i) A 1.6 Km route in Shenzhen, Guangdong for lunch meal delivery. Photos taken in Zibo, Shandong: (b) Starting from the logistics company; (c) Crossing the gate of the logistics company; (d) and (e) Urban main roads. Photos taken in Suzhou, Jiangsu: (g) Left turn; (h) Spraying disinfectant in a residential area. Photos taken in Shenzhen, Guangdong: (j) Heavy traffic environment; (k) Surrounded with road users and meeting with oncoming cars; (l) U-turn in a narrow road; (m) Traffic light at night; (n) Contact-less picking up meal.}
	\label{fig:Impression}
\end{figure*}

\begin{table*}[t]
	\setlength{\abovecaptionskip}{0pt} 
	\setlength{\belowcaptionskip}{0pt} 
	\renewcommand\arraystretch{1.15}   
	\renewcommand\tabcolsep{8pt}  
	\centering
	\caption{The details of the contact-less goods transportation tasks during the COVID-19 pandemic in China.}
	\begin{tabular}{ccccccc}
		\hline
		City                                                                           & Task               & Distance & Time Duration & Payload                                                              & Environments                                                               & Characteristics                                                                    \\ \hline
		\multirow{2}{*}{\begin{tabular}[c]{@{}c@{}}\\Zibo,\\ Shandong\end{tabular}}     & Vegetable Delivery & 9.6 Km   & 75 min        & 600 KG                                                               & Urban Main Road                                                            & \begin{tabular}[c]{@{}c@{}}Light Traffic\\ Heavy Payload\\ Slow Speed\end{tabular} \\ \cline{2-7} 
		& Vegetable Delivery & 5.4 Km   & 50 min        & 960 KG                                                               & \begin{tabular}[c]{@{}c@{}}Urban Main Road\\ Residential Area\end{tabular} & \begin{tabular}[c]{@{}c@{}}Light Traffic\\ Heavy Payload\\ Slow Speed\end{tabular} \\ \hline
		\multirow{2}{*}{\begin{tabular}[c]{@{}c@{}}\\Suzhou,\\ Jiangsu\end{tabular}}     & Meal Delivery      & 1.2 Km   & 30 min        & \begin{tabular}[c]{@{}c@{}}80 KG\\ (100 boxes of meals)\end{tabular} & Urban Main Road                                                            & \begin{tabular}[c]{@{}c@{}}Medium Traffic\\ Left-turn\\ U-turn\end{tabular}        \\ \cline{2-7} 
		& Road Disinfection  & 0.6 Km   & 10 min        & -                                                                    & Residential Area                                                           & \begin{tabular}[c]{@{}c@{}}Narrow Road\\ Crowded Obstacle\\ Slow Speed \end{tabular}             \\ \hline
		\multirow{2}{*}{\begin{tabular}[c]{@{}c@{}}\\Shenzhen,\\ Guangdong\end{tabular}} & Meal Delivery      & 1.6 Km   & 20 min        & \begin{tabular}[c]{@{}c@{}}64 KG\\ (80 boxes of meals)\end{tabular}  & \begin{tabular}[c]{@{}c@{}}Urban Main Road\\ Residential Area\end{tabular} & \begin{tabular}[c]{@{}c@{}}Heavy Traffic\\ Narrow Road\\ U-turn\end{tabular}       \\ \cline{2-7} 
		& Meal Delivery      & 4.0 Km   & 40 min        & \begin{tabular}[c]{@{}c@{}}96 KG\\ (120 boxes of meals)\end{tabular} & \begin{tabular}[c]{@{}c@{}}Urban Main Road\\ Residential Area\end{tabular} & \begin{tabular}[c]{@{}c@{}}Heavy Traffic\\ Narrow Road\\ U-turn\end{tabular}       \\ \hline
	\end{tabular}
	\label{table:Operation record}
\end{table*}

\section{Evaluation} 
 
This section describes the real tasks of contact-less goods transportation using our vehicle during the COVID-19 pandemic in China. From Feb. 2, 2020 to May. 27, 2020, we have deployed 25 vehicles in 3 different cities (Zibo, Shandong; Suzhou, Jiangsu; Shenzhen, Guangdong) in China. Our current server can handle 200 vehicles simultaneously. The total running distances of each vehicle reached 2,500 Km. The details of the transportation tasks are summarized in Tab. \ref{table:Operation record}. Selected demonstration photos during the tasks are displayed in Fig. \ref{fig:Impression}.  
Note that in case of failures during the autonomous navigation, such as unavoidable accidents and system errors, we build a parallel driving system to back up our vehicle control. The parallel driving system is a remote control system based on 4G/5G technology. When using 4G with good signal, the latency is usually between 30ms and 60ms. We set the system to automatically adjust the bit rate to ensure that the vehicle is not out of line. When using 5G, the latency can be less than 20ms. If the vehicle is out of line, it will stop immediately. We have tested the function with several vehicles on several real road environments, and the experimental results show that the function works well. The vehicle control is immediately taken over by a human driver in case of any failure for the autonomous navigation. We expect less human intervention during our goods transportation tasks. The performance is evaluated by the number of occurrences of human interventions.
 
\section{Lessons Learned and Conclusions}
 
For object detection, we found that real-time performance deserves much more attention than accuracy in practice. The perception algorithms should be efficient since they lie in the front-end of the whole autonomous navigation system.  
Therefore, we replaced the dense convolution layers with spatially sparse convolution in our 3-D object detection module. As a result, the inference time is boosted from $250$ ms approximately to $56$ ms.

For the point-cloud mapping, we found that our system was capable of dealing with the rapid motion of the vehicle and short-term point-cloud occlusions. Since most of the Lidars are mounted parallel to the ground and the vehicles always move along the ground, the typical ring structure of the point clouds makes the system difficult to observe in terms of translational movements vertical to the ground plane. Drift in this direction is inevitable during long-term operations. In practice, we learnt that the GPS localization results signalled the potential loop, leading to a consistent larger-region map for Lidar-based localization. In very crowded dynamic traffic environments, the system could be degraded by the disturbances from moving objects \cite{sun2017improving}. To tackle this issue, we use semantic segmentation to remove movable objects to get clear point-cloud data.   

For the planning part, we found that the four-layer hierarchical planning is necessary and effective. The working environments of our vehicle are complex in terms of road structures, traffic conditions, driving etiquette, etc. Layer-wise planning makes the system extensible for multiple environments and various requirements. Furthermore, it is essential to attach importance to the uncertainty from the perception modules. The uncertainty comes from the limited accuracy of the perception system as well as the time delay in processing. For the planning, we avoid hitting the critical conditions and leave safe distances for the planned trajectory.  

For the control part, we found that the greatest advantage of using an MPC can be gained by adding multiple constraints in the control process. When the vehicle operates at low speed, the kinematic constraints restrain the vehicle motion planning and control. But with the increment of speed, dynamic characteristics become more influential. As aforementioned, the dynamic-based MPC is much more accurate than the kinematic-based MPC since the predictive model is more accurate. However, we found that the complex model prediction is not the best option. With regards to low- and middle-speed operation environments, a simplified predictive model with multiple constraints would be sufficient. 
 
From the real-life operations, we found that more conservative settings for obstacle detection could lead to more false positives. This would decrease the speed or even freeze the vehicle, and hence cause traffic jams. On some roads, the minimum allowed speed is not indicated, we need to keep the vehicle speed not too slow. Otherwise, it would cause annoyance for other vehicle drivers. 
Clearly and easily identified human-machine interface is also important. It can be used to inform other vehicle drivers in advance what the autonomous vehicle will do. Otherwise, the other vehicle drivers would feel frightened because they could not anticipate the behaviours of the autonomous vehicle. For example, our vehicle often frightens other vehicle drivers when it is making a reverse even the reversing light is flashing. Using a screen to notify the reversing behaviour could alleviate the issue.
In some cases, not strictly obeying the traffic rules would be good for the autonomous navigation. For example, it would be wise to change the lane when there happens a traffic accident ahead of the ego-lane, even the lane changing behaviour is not allowed according to the traffic rules. Otherwise, the vehicle could not get over.

The successful daily operations demonstrated that using our autonomous logistic vehicle could effectively avoid virus spread due to human contact. It effectively builds a virtual wall between the recipient and sender during the goods transportation. 
For quantitative measures, we can compute from Tab. \ref{table:Operation record} that the average distance for each task per vehicle is $(9.6+5.4+1.2+0.6+1.6+4.0)/6\approx3.7$ Km. As the total running distance is 2,500 Km, the number of tasks is $2,500/3.7\approx676$. According to our observation, there are usually $4$ times of person-to-person contacts in each task of the traditional goods transportation. So the number of avoided contacts would be $4\times676=2,704$. As we have 25 running vehicles, the total number of avoided contacts would be $25\times2,704=67,600$. 
Currently, there is a huge demand for contact-less goods transportation in many infected areas. We believe that continuous long-term operations could extensively improve our vehicle and enhance the maturity of our technologies.
 
\section{Acknowledgements}
This work was supported by the Hong Kong RGC Project No. 11210017, Guangdong Science and Technology Plan Guangdong-Hong Kong Cooperation Innovation Platform Project No. 2018B050502009, Shenzhen Science and Technology Innovation Commission Project No. JCYJ2017081853518789, Macao Science and Technology Development Fund Project No. 0015/2019/AKP. 
 
\addtolength{\textheight}{0cm}   
 
\bibliographystyle{IEEEtran}
\bibliography{IEEEabrv,reference}

\vspace{0.5cm}  

\textbf{Tianyu Liu} Shenzhen Unity Drive Innovation Technology Co. Ltd., Shenzhen, China. Email: liutianyu@unity-drive.com.

\textbf{Qinghai Liao} The Hong Kong University of Science and Technology, Clear Water Bay, Hong Kong, China. Email: qinghai.liao@connect.ust.hk.

\textbf{Lu Gan} The Hong Kong University of Science and Technology, Clear Water Bay, Hong Kong, China. Email: lganaa@connect.ust.hk.

\textbf{Fulong Ma} The Hong Kong University of Science and Technology, Clear Water Bay, Hong Kong, China. Email: fmaaf@connect.ust.hk.

\textbf{Jie Cheng} The Hong Kong University of Science and Technology, Clear Water Bay, Hong Kong, China. Email: jchengai@connect.ust.hk.

\textbf{Xupeng Xie} The Hong Kong University of Science and Technology, Clear Water Bay, Hong Kong, China. Email: xxieak@connect.ust.hk.

\textbf{Zhe Wang} Shenzhen Unity Drive Innovation Technology Co. Ltd., Shenzhen, China. Email: wangzhe@unity-drive.com.

\textbf{Yingbing Chen} The Hong Kong University of Science and Technology, Clear Water Bay, Hong Kong, China. Email: ychengz@connect.ust.hk.

\textbf{Yilong Zhu} The Hong Kong University of Science and Technology, Clear Water Bay, Hong Kong, China. Email: yzhubr@connect.ust.hk.

\textbf{Shuyang Zhang} Shenzhen Unity Drive Innovation Technology Co. Ltd., Shenzhen, China. Email: shuyang.zhang@unity-drive.com.

\textbf{Zhengyong Chen} Shenzhen Unity Drive Innovation Technology Co. Ltd., Shenzhen, China. Email: chenzhengyong@unity-drive.com.

\textbf{Yang Liu} The Hong Kong University of Science and Technology, Clear Water Bay, Hong Kong, China. Email: yang.liu@connect.ust.hk.

\textbf{Meng Xie} Shenzhen Unity Drive Innovation Technology Co. Ltd., Shenzhen, China. Email: xiemeng@unity-drive.com.

\textbf{Yang Yu} The Hong Kong University of Science and Technology, Clear Water Bay, Hong Kong, China. Email: yyubj@connect.ust.hk.

\textbf{Zitong Guo} Shenzhen Unity Drive Innovation Technology Co. Ltd., Shenzhen, China. Email: guozitong@unity-drive.com.

\textbf{Guang Li} Shenzhen Unity Drive Innovation Technology Co. Ltd., Shenzhen, China. Email: liguang@unity-drive.com.

\textbf{Peidong Yuan} Shenzhen Unity Drive Innovation Technology Co. Ltd., Shenzhen, China. Email: yuanpeidong@unity-drive.com.

\textbf{Dong Han} Shenzhen Institutes of Advanced Technology, Chinese Academy of Sciences, Shenzhen, China. Email: dong.han@siat.ac.cn.

\textbf{Yuying Chen} The Hong Kong University of Science and Technology, Clear Water Bay, Hong Kong, China. Email: ychenco@connect.ust.hk.

\textbf{Haoyang Ye} The Hong Kong University of Science and Technology, Clear Water Bay, Hong Kong, China. Email: hy.ye@connect.ust.hk.

\textbf{Jianhao Jiao} The Hong Kong University of Science and Technology, Clear Water Bay, Hong Kong, China. Email: jjiao@connect.ust.hk.

\textbf{Peng Yun} The Hong Kong University of Science and Technology, Clear Water Bay, Hong Kong, China. Email: pyun@connect.ust.hk.

\textbf{Zhenhua Xu} The Hong Kong University of Science and Technology, Clear Water Bay, Hong Kong, China. Email: zxubg@connect.ust.hk.

\textbf{Hengli Wang} The Hong Kong University of Science and Technology, Clear Water Bay, Hong Kong, China. Email: hwangdf@connect.ust.hk.

\textbf{Huaiyang Huang} The Hong Kong University of Science and Technology, Clear Water Bay, Hong Kong, China. Email: hhuangat@connect.ust.hk.

\textbf{Sukai Wang} The Hong Kong University of Science and Technology, Clear Water Bay, Hong Kong, China. Email: swangcy@connect.ust.hk.

\textbf{Peide Cai} The Hong Kong University of Science and Technology, Clear Water Bay, Hong Kong, China. Email: peide.cai@connect.ust.hk.

\textbf{Yuxiang Sun} The Hong Kong Polytechnic University, Hung Hom, Kowloon, Hong Kong. Email: yx.sun@polyu.edu.hk.

\textbf{Yandong Liu} Shenzhen Institutes of Advanced Technology, Chinese Academy of Sciences, Shenzhen, China. Email: yd.liu@siat.ac.cn.

\textbf{Lujia Wang} Shenzhen Institutes of Advanced Technology, Chinese Academy of Sciences, Shenzhen, China. Email: lj.wang1@siat.ac.cn.

\textbf{Ming Liu} The Hong Kong University of Science and Technology, Clear Water Bay, Hong Kong, China. Email: eelium@ust.hk.

\end{document}